\documentclass[11pt]{article}

\usepackage[final]{acl}

\usepackage{times}
\usepackage{latexsym}
\usepackage[T1]{fontenc}
\usepackage[utf8]{inputenc}
\usepackage{microtype}
\usepackage{inconsolata}
\usepackage{amsmath,amsfonts,amssymb}
\usepackage{booktabs}
\usepackage{graphicx}
\usepackage{tikz}
\usetikzlibrary{arrows.meta,positioning,fit,shapes.geometric}
\usepackage{algorithm}
\usepackage{algorithmic}

\title{DQA: Diagnostic Question Answering for IT Support}

\author{Vishaal Kapoor \quad Mariam Dundua \quad Sarthak Ahuja \quad Neda Kordjazi \quad Evren Yortucboylu \\
  {\bf Vaibhavi Padala} \quad {\bf Derek Ho} \quad {\bf Jennifer Whitted} \quad {\bf Rebecca Steinert} \\
  Amazon \\
  {\small \texttt{\{vishaalk,madundua,sarahuja,nedakord,yortuc,vapadala,derekjho,jenwhitt,rsteinrt\}@amazon.com}}}

\begin{document}
\maketitle

\begin{abstract}
Enterprise IT support interactions are fundamentally diagnostic: effective resolution requires iterative evidence gathering from ambiguous user reports to identify an underlying root cause. While retrieval-augmented generation (RAG) provides grounding through historical cases, standard multi-turn RAG systems lack explicit diagnostic state and therefore struggle to accumulate evidence and resolve competing hypotheses across turns.

We introduce DQA, a diagnostic question-answering framework that maintains persistent diagnostic state and aggregates retrieved cases at the level of root causes rather than individual documents. DQA combines conversational query rewriting, retrieval aggregation, and state-conditioned response generation to support systematic troubleshooting under enterprise latency and context constraints.

We evaluate DQA on 150 anonymized enterprise IT support scenarios using a replay-based protocol. Averaged over three independent runs, DQA achieves a 78.7\% success rate under a trajectory-level success criterion, compared to 41.3\% for a multi-turn RAG baseline, while reducing average turns from 8.4 to 3.9. This improvement reflects the benefit of explicitly representing competing explanations and aggregating evidence across turns in unscripted troubleshooting.

\end{abstract}

\section{Introduction}

Within enterprise IT support environments, interactions are fundamentally diagnostic: users present incomplete and ambiguous symptoms, and support agents iteratively gather evidence to identify an underlying root cause. Unlike single-turn fact-based question answering, effective troubleshooting requires tracking competing hypotheses, interpreting partial signals, and deciding when to ask clarifying questions versus proposing resolutions. These properties make enterprise IT support a challenging setting for retrieval-augmented language models.

Retrieval-augmented generation (RAG) is a standard approach for grounding language models in external evidence. Extensions to multi-turn settings incorporate conversational query rewriting to improve retrieval under contextual references.

However, standard multi-turn RAG systems typically lack an explicit representation of diagnostic state. Retrieved documents are consumed independently at each turn, making it difficult to accumulate evidence across the interaction, reconcile conflicting signals, or maintain awareness of unresolved hypotheses. As a result, conversational coherence is often conflated with diagnostic progress.

This setting raises three challenges: limited context, lack of diagnostic state, and principled action selection under uncertainty.

We introduce DQA, a diagnostic question-answering framework that explicitly maintains diagnostic state (relative support over competing root-cause hypotheses and accumulated evidence) across turns and aggregates retrieved evidence at the level of root causes rather than individual documents. DQA is designed to support systematic troubleshooting under strict latency and context constraints typical of enterprise environments.

DQA aggregates retrieved cases into root-cause--level signals, maintains a diagnostic belief state tracking competing hypotheses, and uses this state to guide questioning and resolution. This separation of evidence aggregation from decision-making enables multi-turn diagnostic reasoning without expanding the context window.

Our contribution is a system-level design that integrates retrieval aggregation, persistent diagnostic state, and state-conditioned action selection for enterprise IT support. While evaluated in an IT support setting, the design principles underlying DQA apply more broadly to retrieval-augmented systems that must reason over large case repositories under bounded context and latency.

\section{Background and Related Work}

Retrieval-augmented generation (RAG) conditions language model outputs on retrieved external evidence to improve factual grounding~\citep{lewis2020rag}. In enterprise IT support, the retrieved corpus often consists of internal documentation and large ticket repositories. At this scale, raw ticket retrieval yields redundant near-duplicate evidence: many retrieved cases describe the same symptoms and resolutions with superficial variation, wasting context and latency budget. Case-based reasoning similarly frames troubleshooting as retrieving similar resolved cases followed by adaptation~\citep{aamodt1994case}; DQA follows this spirit but targets aggregation—compressing large retrieved neighborhoods into compact diagnostic signals rather than adapting from a few exemplar cases. Whereas simple deduplication removes redundancy by collapsing similar cases, aggregation preserves distributional information (e.g., cluster prevalence) that guides downstream action selection.

\textbf{Multi-turn conversational retrieval.}
Multi-turn retrieval is difficult because user turns are frequently non-standalone (e.g., anaphora, ellipsis, follow-ups), causing retrieval drift unless context is resolved~\citep{choi2018quac,reddy2019coqa}. Conversational query rewriting (CQR) rewrites each user turn into a standalone query to improve retrieval robustness under contextual references~\citep{yu2020convqrewriting,vakulenko2021qrcompare,qian2022explicitqr}. Benchmarks such as MTRAG and CORAL show that conversational RAG quality degrades over longer interactions and that rewriting can mitigate retrieval failures~\citep{katsis2025mtrag,cheng2025coral}. However, even with strong rewriting, these systems do not explicitly represent diagnostic progress or track which explanations remain plausible as evidence accumulates.

\textbf{Diagnostic dialogue.}
Diagnostic assistance differs from fact-seeking QA: the assistant must choose questions and actions that reduce uncertainty over a latent cause rather than directly returning an answer. Prior diagnostic systems in technical and medical domains often rely on decision trees or manually engineered symptom hierarchies, which can be brittle and costly to maintain as environments evolve~\citep{aamodt1994case}. Enterprise IT support amplifies these issues because user descriptions are noisy, infrastructures are heterogeneous, and failure modes shift over time, making static workflows difficult to sustain.

\textbf{Gap addressed by DQA.}
Prior multi-turn RAG systems improve retrieval robustness but do not explicitly represent evolving diagnostic context—which hypotheses remain plausible or how evidence has shifted across turns. DQA introduces persistent diagnostic state and evidence aggregation to enable systematic uncertainty reduction in unscripted troubleshooting.

\section{Diagnostic Question Answering (DQA)}

We introduce Diagnostic Question Answering (DQA), a framework for multi-turn enterprise IT support that treats troubleshooting as progressive uncertainty reduction over latent root causes. DQA operates in unscripted diagnostic settings and maintains explicit diagnostic state across turns, enabling systematic action selection under uncertainty.

\subsection{Evidence Aggregation}

DQA constructs data-driven priors over plausible root causes using historical support tickets. Each resolved ticket contains a free-text resolution field extracted from the interaction transcript, which we treat as a proxy for the underlying root cause.

Given an initial user description, the system retrieves similar tickets and clusters them by their resolution embeddings. The clusters represented in this neighborhood define the candidate hypotheses, and their relative frequencies form an empirical prior. This aggregation deduplicates near-identical cases and combines evidence across many similar incidents, producing a compact diagnostic signal.

\subsection{Retrieval-Induced Diagnostic State}

DQA maintains diagnostic state using a structured representation that summarizes support for each candidate root cause. The state includes a hypothesis-weight vector $\mathbf{h}_t \in \mathbb{R}^K$ (where each element corresponds to a root-cause cluster), along with associated evidence such as representative symptoms, KB articles, and canonical resolutions.

As new information becomes available, the system re-retrieves and re-aggregates evidence, updating the diagnostic state without requiring explicit symbolic belief tracking or hand-engineered probability models. Structured state fields persist across turns, while retrieval-induced weights are recomputed from freshly aggregated evidence.

\subsection{Action-Aware Diagnostic Policy}

The diagnostic state guides action selection. Rather than generating unconstrained free-form responses, DQA frames troubleshooting as a policy over a small set of diagnostic actions~\citep[cf.][]{yao2023react}. This abstraction makes diagnostic progress explicit by separating evidence formation from decisions about what to do next.

Concretely, the policy operates over three action types: clarifying questions, investigative steps, and resolution proposals. These correspond to gathering discriminative evidence, validating likely causes, and proposing a fix once uncertainty is reduced.

As evidence accumulates and support concentrates on a few root causes, the policy shifts from broad questioning toward targeted investigation and resolution. Section 5.2 describes how this policy is executed via state-conditioned response generation.

\section{RAggG: Retrieval-Aggregated Generation for Root-Cause Priors}

We now describe RAggG (Retrieval-Aggregated Generation) in detail. Standard RAG retrieves individual documents and conditions generation on raw retrieved text. When applied to large repositories of historical support cases, this leads to redundancy and poor signal-to-noise ratios: many retrieved tickets describe near-identical symptoms and resolutions with only superficial variation. RAggG addresses this by aggregating retrieved evidence along a task-relevant dimension---root cause---prior to generation.

RAggG operates in three stages: (1) retrieval selects a large candidate neighborhood, (2) aggregation clusters candidates by root-cause descriptions and computes per-cluster evidence counts and representative cases, and (3) generation operates on these aggregates rather than individual tickets. Algorithm~\ref{alg:ragg} summarizes the approach.

DQA operates as a multi-turn control loop. At each turn, the current diagnostic state conditions query rewriting, retrieval, and action selection. RAggG is used as an aggregation subroutine to compress the retrieved neighborhood into updated diagnostic evidence. We maintain a structured diagnostic state $s_t$ across turns, which includes a retrieval-induced weight vector $\mathbf{h}_t \in \mathbb{R}^K$ as one component, along with additional fields such as extracted symptoms, exemplar tickets, and canonical resolutions.

\begin{algorithm}[t]
\small
\caption{RAggG: Retrieval Neighborhood Aggregation}
\label{alg:ragg}
\begin{algorithmic}[1]
\STATE \textbf{Input:} User query $x$, ticket repository $\mathcal{D}$
\STATE Encode query: $z_x \gets f(x)$
\STATE Retrieve similar tickets: $\mathcal{N}(x) \gets \mathrm{TopK}(\mathcal{D}, z_x, K)$
\STATE Cluster retrieved tickets on root-cause:
\STATE \hspace{1em}$\{C_j\}_{j=1}^{J} \gets \mathrm{Cluster}(\mathcal{N}(x))$
\FOR{each cluster $C_j$}
    \STATE Count evidence mass: $n_j \gets |C_j|$ \COMMENT{raw evidence counts}
    \STATE Select representative cases: $R_j \gets \mathrm{SelectRepr}(C_j)$
\ENDFOR
\STATE \textbf{Output:} Aggregated evidence $\mathcal{E} = \{(n_j, R_j)\}_{j=1}^{J}$
\end{algorithmic}
\end{algorithm}

\begin{algorithm}[t]
\small
\caption{DQA: Diagnostic State Update (Event-Driven)}
\label{alg:dqa}
\begin{algorithmic}[1]
\STATE \textbf{Input:} User query $x$, conversation history $\mathcal{H}$, diagnostic state $s$, repository $\mathcal{D}$
\STATE Rewrite query: $\tilde{x} \gets \mathrm{Rewrite}(x, \mathcal{H}, s)$
\STATE Aggregate evidence (Algorithm~\ref{alg:ragg}):
\STATE \hspace{1em}$\mathcal{E} \gets \textsc{RAggG}(\tilde{x}, \mathcal{D})$
\STATE Update diagnostic state schema/content:
\STATE \hspace{1em}$s \gets \mathrm{UpdateState}(s, \mathcal{E}, \mathcal{H})$
\STATE Compute retrieval-induced weights (from counts in $\mathcal{E}$):
\FOR{each $(n_j, \cdot) \in \mathcal{E}$}
    \STATE $h_j \gets \dfrac{n_j}{\sum_{(n_{j'}, \cdot)\in \mathcal{E}} n_{j'}}$
\ENDFOR
\STATE Store retrieval-conditioned hypothesis state in $s$: $s.h \gets h$
\STATE Generate response:
\STATE \hspace{1em}$y \gets \mathrm{GenerateResponse}(\mathcal{H} \cup \{x\}, s, \mathcal{E})$
\STATE Update history: $\mathcal{H} \gets \mathcal{H} \cup \{(x, y)\}$
\STATE \textbf{Output:} Response $y$
\end{algorithmic}
\end{algorithm}

State updates are implemented via re-retrieval and re-aggregation at each turn, rather than explicitly propagating probabilities over $\mathbf{h}_t$. This design recomputes retrieval-induced weights from fresh evidence while maintaining a persistent structured diagnostic state.

\subsection{Ticket Representation and Clustering}
Each resolved ticket $d \in \mathcal{D}$ is processed with an encoder that extracts: (i) a short free-text \emph{root cause description} (resolution field extracted from the interaction transcript), (ii) a summary of user-reported and engineer-confirmed symptoms, and (iii) resolution steps.
We encode the root cause description with a sentence encoder $f(\cdot)$~\citep{reimers2019sentencebert} to obtain an embedding $z_d \in \mathbb{R}^h$. We then cluster $\{z_d\}$ in this space using a scalable clustering algorithm (e.g., mini-batch $k$-means or hierarchical agglomerative clustering), yielding clusters $\{C_1, \ldots, C_K\}$.

Each cluster $C_k$ is interpreted as a data-driven approximation to a latent root cause $r_k$, with an empirical prior
\begin{equation}
    p(r_k) = \frac{|C_k|}{|\mathcal{D}|},
\end{equation}
which is reported for interpretability only and not used online.

\subsection{Query-Conditioned Priors and Scalability}
Given a new contact with description $x$, we compute a query embedding $z_x = f(x)$ and retrieve the top-$K$ nearest tickets in embedding space. Let $\mathcal{N}(x)$ denote this neighborhood and let $n_k(x)$ be the number of retrieved tickets that fall in cluster $C_k$. At interaction time, DQA defines a retrieval-conditioned hypothesis distribution over the clusters present in the neighborhood:
\begin{equation}
    h_k = \frac{n_k(x)}{\sum_{k' \in K(x)} n_{k'}(x)},
\end{equation}
where $K(x)$ is the set of clusters that appear in the retrieved neighborhood.

The aggregation step is crucial for scalability. Rather than passing hundreds of individual tickets to downstream processing, RAggG operates on a small number of aggregate statistics (typically $\leq$5 clusters), enabling real-time performance even with repositories containing 100K+ historical cases.

\section{Diagnostic State and Action Selection}

The diagnostic state is represented as a structured state object that includes a retrieval-induced weight vector $\mathbf{h} \in \mathbb{R}^K$, where $\sum_k h_k = 1$, with each element corresponding to a root-cause cluster induced from historical tickets. Each entry $h_k$ represents the relative prevalence of cluster $C_k$ within the current retrieval neighborhood, computed by normalizing evidence counts obtained via aggregation. The weight vector is recomputed at each turn via re-retrieval and aggregation rather than explicitly propagating probabilities across turns.

\subsection{Diagnostic State Schema}
At each turn, the diagnostic state maintains four components: (i) a current hypothesis string summarizing the working diagnostic theory, (ii) a list of user-reported symptoms extracted from the conversation, (iii) a ranked list of knowledge base (KB) articles, dynamically filtered and re-ranked based on the current diagnostic hypothesis state, and (iv) a set of candidate root-cause clusters, each associated with a retrieval-induced weight $h_k$, a root-cause description, and representative resolved tickets.

State updates are hybrid. Structured, non-weight fields of the diagnostic state—such as the hypothesis string, symptom list, knowledge base references, and the candidate set of root-cause clusters—are updated incrementally as new information becomes available. Retrieval-induced weights are then recomputed from the current aggregated evidence and stored in the state, rather than numerically propagated from the previous turn.

This structured diagnostic state is serialized into a prompt that the language model conditions on when selecting the next diagnostic action.

\subsection{Action Selection}

DQA implements the policy from Section 3.3 through state-conditioned response generation. At each turn, the language model is provided with the current diagnostic state (including the leading hypotheses and accumulated evidence) together with the conversation history and aggregated retrieval summaries, and it generates the next response accordingly. A single generated response may incorporate elements of multiple action types (e.g., proposing a targeted investigative step while asking a clarifying question).

Conditioning on explicit diagnostic state encourages discriminative next steps rather than generic information requests.

\subsection{Query Rewriting for Conversational Retrieval}

Multi-turn enterprise support interactions frequently contain anaphora, informal terminology, and underspecified follow-ups (e.g., "still broken"), which can cause retrieval drift if treated in isolation. We apply lightweight conversational query rewriting to normalize terminology and resolve references.

The rewriter conditions on both dialogue history and the current diagnostic state. When users provide vague follow-ups after a diagnostic hypothesis has emerged, the rewriter incorporates the dominant diagnostic context, maintaining focus on plausible root causes and preventing retrieval from reverting to generic failure modes.

\section{Experimental Setup}

We evaluate DQA using a replay-based protocol on real enterprise IT support interactions. This section describes the dataset, evaluation procedure, system variants, and metrics.

\subsection{Dataset and Replay Protocol}

We evaluate on 150 anonymized enterprise IT support interactions, each consisting of a multi-turn interaction, labeled with a final resolved root cause. The evaluation set is intentionally heterogeneous, spanning hardware failures, software configuration issues, access management, authentication systems, and enterprise applications, with a majority of cases involving permission and access-related failures, reflecting the distribution observed in real-world enterprise helpdesk workloads.

To enable controlled comparison under realistic diagnostic conditions, we adopt a replay-based evaluation protocol. At each replay turn $t$, systems are provided with the dialogue prefix up to that point and generate a response, which may emphasize information gathering, investigative steps, or resolution proposals. The interaction ends when the system proposes a resolution or after 15 turns.

User responses are generated by an LLM simulating the customer, conditioned on a scenario-specific persona (communication style, knowledge level, and priorities) and deterministic state transitions that define outcomes of investigative actions (e.g., "restart performed $\rightarrow$ same issue persists"). Trajectories diverge across systems: each system's responses produce different simulated user reactions, enabling evaluation of diagnostic steering rather than response quality to fixed inputs.

System responses are generated offline to produce complete interaction transcripts. A structured post-hoc extraction pass identifies diagnostic facts, resolution steps, and antipattern violations for evaluation. All systems share the same dialogue context, evidence sources, and retrieval infrastructure; differences arise from how evidence is formulated.

\subsection{Systems Compared and Ablations}

We compare DQA against increasingly capable RAG-based systems that isolate the effects of individual components. RAG (no CQR) uses raw user utterances for retrieval without conversational rewriting. RAG (baseline) adds query rewriting to produce standalone retrieval queries. RAG + Clustering extends the baseline by aggregating retrieved tickets via root-cause clustering (RAggG), without maintaining persistent diagnostic state. DQA further adds a persistent diagnostic state to guide action selection across turns. All systems share the same retrieval infrastructure, encoders, language model, and high-level prompting framework, operating over a combined corpus of historical tickets and KB articles; aggregation and diagnostic hypothesis induction are applied to tickets, while KB articles are surfaced as supporting evidence. DQA additionally conditions responses on explicit diagnostic state maintained across turns to guide action selection.

\subsection{Evaluation Metrics}

We evaluate systems using a task-level success criterion aligned with enterprise troubleshooting requirements. Each scenario specifies a set of required diagnostic facts, required resolution steps, and one antipattern constraint. All three components are evaluated by an LLM judge~\citep{zheng2023judging} applied to the full interaction trajectory.

Diagnosis score measures the fraction of required diagnostic facts correctly identified (range 0--1). Resolution score measures the fraction of required resolution steps provided (range 0--1). Antipattern compliance evaluates whether the agent exhibited a scenario-specific problematic behavior—such as redundant troubleshooting of steps the user has already attempted, pursuing a diagnostic path that contradicts available symptoms, or setting false expectations about resolution timelines.

A test case is considered successful if, at any point during the interaction, the system achieves diagnosis score = 1.0, resolution score = 1.0, and satisfies the antipattern constraint. Success does not require the system to terminate on the turn at which these criteria are met. Scores reported in Table 1 are averaged over all 150 scenarios, including unsuccessful cases.

We additionally report average interaction length (number of system turns until termination) as a secondary efficiency metric.

\section{Results}

This section reports comparative performance between DQA and RAG-based baselines under the replay-based evaluation protocol described in Section 6. All reported results are averaged over three independent executions of the replay protocol.

\subsection{Overall Diagnostic Performance}

Table~\ref{tab:main_results} summarizes diagnostic performance across system variants under the success criterion defined in Section 6.3. The full DQA system achieves a 78.7\% success rate, outperforming all ablated variants under the replay-based protocol.

Compared to the baseline RAG system (41.3\%), DQA improves success rate by 37.4 percentage points, a 90.6\% relative improvement. Improvements are observed across diagnosis completeness, resolution completeness, and antipattern avoidance.

These gains are consistent with the hypothesis that persistent diagnostic state enables evidence accumulation across turns. In addition to higher diagnostic success, DQA converges substantially faster. Averaged over three independent runs, DQA resolves successful cases in 3.9 turns on average, compared to 8.4 turns for the baseline RAG system.

\begin{table}[t]
\small
\centering
\begin{tabular}{lcccc}
\toprule
\textbf{System} & \textbf{Success} & \textbf{Diag.} & \textbf{Res.} & \textbf{Turns $\downarrow$} \\
\midrule
RAG (no CQR) & 40.4\% & 0.86 & 0.73 & 8.17 \\
RAG (baseline) & 41.3\% & 0.86 & 0.73 & 8.43 \\
RAG + Clustering & 53.8\% & 0.90 & 0.80 & 6.53 \\
DQA (ours) & \textbf{78.7\%} & \textbf{0.99} & \textbf{0.94} & \textbf{3.93} \\
\midrule
$\Delta$ vs baseline & \textbf{+37.4pp} & \textbf{+0.13} & \textbf{+0.21} & \textbf{$-$4.50} \\
\bottomrule
\end{tabular}
\caption{Diagnostic performance averaged over three runs on 150 enterprise IT support scenarios. DQA improves both end-to-end success and convergence efficiency, resolving cases in substantially fewer system interaction turns than RAG-based baselines.}
\label{tab:main_results}
\end{table}

\subsection{Ablation Results}

Ablation results in Table~\ref{tab:ablation} isolate the contributions of individual system components. Conversational query rewriting provides a modest improvement over raw retrieval, primarily by mitigating retrieval drift in context-dependent user utterances. The relatively modest gains from conversational query rewriting likely reflect the evaluation setup: replayed user turns, including simulated persona variations, are generally more well-formed and information-dense than many live enterprise inputs, which are often underspecified or noisy—conditions under which CQR is known to provide larger benefits. Retrieval aggregation via clustering improves success rate from 41.3\% to 53.8\%, indicating that compressing retrieved evidence at the root-cause level improves diagnostic signal quality.

Introducing persistent diagnostic state further increases success to 78.7\%, representing the largest incremental gain among the evaluated components. These results suggest that persistent diagnostic state contributes more to end-to-end troubleshooting success than retrieval improvements alone.

\begin{table}[t]
\small
\centering
\begin{tabular}{lcc}
\toprule
\textbf{Component Introduced} & \textbf{Success $\Delta$} & \textbf{Relative Gain} \\
\midrule
Query Rewriting (CQR) & +0.9pp & +2.2\% \\
Semantic Clustering & +12.5pp & +30.3\% \\
Diagnostic State & +24.9pp & +46.3\% \\
\bottomrule
\end{tabular}
\caption{Ablation results showing the effect of query rewriting, retrieval clustering, and persistent diagnostic state.}
\label{tab:ablation}
\end{table}

\subsection{Latency and Scalability}

DQA incurs one additional LLM call per turn for query rewriting and hypothesis-conditioned state updates, but converges in substantially fewer turns (3.9 vs. 8.4), yielding comparable or lower total interaction cost. By structuring interactions into targeted diagnostic actions, DQA focuses computation on uncertainty reduction. Retrieval, clustering, and aggregation use non-LLM preprocessing with offline-cached embeddings, maintaining interactive latency.

\section{Conclusion}

We presented DQA, a framework for multi-turn enterprise IT support that treats troubleshooting as an iterative diagnostic process. DQA integrates RAggG for aggregating large retrieval neighborhoods into root-cause–level evidence, a persistent diagnostic state that organizes evidence across turns, and state-conditioned action selection for questioning, investigation, and resolution.

Rather than propagating probabilistic beliefs, DQA recomputes diagnostic support at each turn from freshly aggregated retrieval evidence while maintaining structured diagnostic context.

More broadly, RAggG illustrates a general pattern for retrieval-augmented systems: compressing large, redundant retrieval results into compact, task-relevant aggregates that can guide multi-turn decision making under latency and context constraints.

\section{Limitations}

This work has several limitations. First, our evaluation uses simulated users rather than live deployment, which enables controlled comparison but does not capture user experience factors such as perceived helpfulness or trust. Second, poorly documented resolutions may yield noisy root-cause structure. Third, we evaluate on a single enterprise IT support domain; generalization to other diagnostic settings (e.g., medical diagnosis, hardware troubleshooting) remains to be demonstrated.

\section*{Ethics Statement}
This work uses anonymized enterprise IT support logs with all personally identifiable information removed.

\bibliography{custom}

\end{document}